\documentclass[conference]{IEEEtran}
\usepackage{dblfloatfix}

\usepackage{cite}

\ifCLASSINFOpdf
   \usepackage[pdftex]{graphicx}
\else
\fi

\usepackage{array}
\newcolumntype{P}[1]{>{\centering\arraybackslash}p{#1}}
\newcolumntype{M}[1]{>{\centering\arraybackslash}m{#1}}

\usepackage{fixltx2e}
\makeatletter
\def\ps@IEEEtitlepagestyle{%
	\def\@oddfoot{\mycopyrightnotice}%
	\def\@evenfoot{}%
}
\def\mycopyrightnotice{%
	{\footnotesize 978-1-5090-6004-7/17/\$31.00 \textcopyright 2017 IEEE\hfill}
	\gdef\mycopyrightnotice{}
}

\begin{document}

%
\title{Handwritten Arabic Numeral Recognition using Deep Learning Neural Networks}


\author{\IEEEauthorblockN{Akm Ashiquzzaman and
		Abdul Kawsar Tushar
	}
	\IEEEauthorblockA{Computer Science and Engineering Department, University of Asia Pacific, Dhaka, Bangladesh\\\{zamanashiq3, tushar.kawsar\}@gmail.com}
}


%


\maketitle

\begin{abstract}
Handwritten character recognition is an active area of research with applications in numerous fields. Past and recent works in this field have concentrated on various languages. Arabic is one language where the scope of research is still widespread, with it being one of the most popular languages in the world and being syntactically different from other major languages. Das et al. \cite{DBLP:journals/corr/abs-1003-1891} has pioneered the research for handwritten digit recognition in Arabic. In this paper, we propose a novel algorithm based on deep learning neural networks using appropriate activation function and regularization layer, which shows significantly improved accuracy compared to the existing Arabic numeral recognition methods. The proposed model gives 97.4 percent accuracy, which is the recorded highest accuracy of the dataset used in the experiment. We also propose a modification of the method described in \cite{DBLP:journals/corr/abs-1003-1891}, where our method scores identical accuracy as that of \cite{DBLP:journals/corr/abs-1003-1891}, with the value of 93.8 percent. 

\end{abstract}


\textbf{Keywords - Deep learning, ConvNet, Handwritten digit recognition, Arabic numeral.}

%
\IEEEpeerreviewmaketitle

\section{Introduction}

The optical character recognition (OCR) of digits on scanned images hold widespread commercial and pedagogical importance in fields such as automatic recognition, check reading, data collection from forms, and textbook digitization. Being still and active area of research, OCR recognition is being acquainted to different languages around the world.

\begin{figure}[!t]
	\centering{\includegraphics[width=80mm]{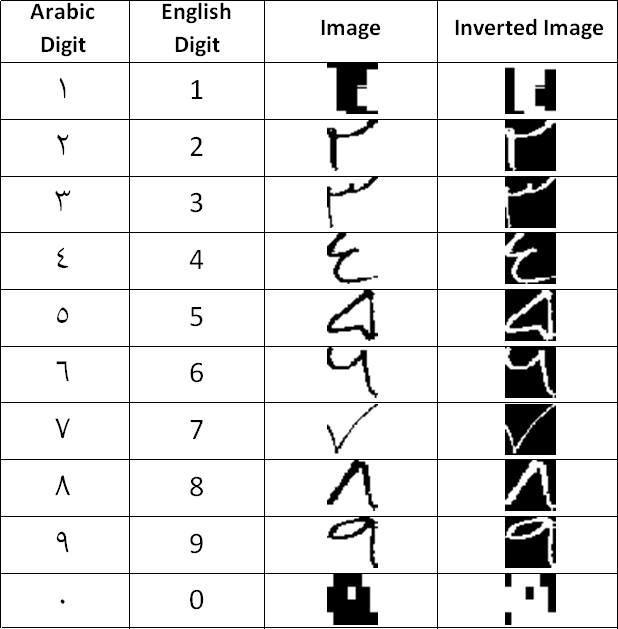}}
	\caption{Arabic handwritten digits and their inverted images.}
	\label{fig:table}
\end{figure}

Statistically speaking, Arabic is one of the top five most widely spoken language in the present world, being used by more than 267 million people \cite{web1}. Arabic is an official language in more than 16 countries and spoken widely in a number of other countries as well \cite{web2}. In recognition of this, Arabic has held the status of an official language of the United Nations since 1973 \cite{web3}. In addition to that, Arabic is the liturgical language of the religion of Islam, hence a large number of religious texts and writings are in Arabic. Numerous languages including Persian, Hindi, and Urdu have taken inspiration from Arabic. Sample handwritten numerals in Arabic language and their inverted images are shown in Fig. \ref{fig:table}.

Although a number of languages have received significant attention in terms of OCR research \cite{plamondon2000online, wong1998off}, Arabic is still mostly an unexplored field of study, with the maximum work done  previously on this language being based on printed characters \cite{amin1998off}. The important distinction between Arabic and major roman-based languages is that Arabic words as well as characters within words are written from right to left, as opposed to left to right as with English. Nonetheless, digits of an Arabic number is written from left to right. 

Das et al. \cite{DBLP:journals/corr/abs-1003-1891} have devised a novel method for Arabic handwritten digit recognition with help of a multi layer perceptron (MLP) which can bring significant accuracy to the method of Arabic digit OCR. In this paper, we propose a methods which can increase the accuracy of this process by using convolutional neural network (CNN) in place of MLP. Furthermore, we also provide a modification of the method of Das et al. which achieves the same level of classification accuracy. Our methods also solve the problem of overfitting that has affected the method in \cite{DBLP:journals/corr/abs-1003-1891}, by applying dropout regularization.

\section{Background}\label{background}

A novel method for recognizing Arabic numerals has been devised by  \cite{DBLP:journals/corr/abs-1003-1891}. This method utilizes an MLP where a set of 88 features is used. The feature set is divided into 72 shadow features and 16 octant features. For this work, a database of 3000 samples is used which is obtained from \cite{web4}. The images are scaled to the pixel size of 32x32 and then defined with gray scale pixel values. Finally, conversion to binary images takes place. 

\begin{figure}[bh]
	\centerline{\includegraphics[width=80mm]{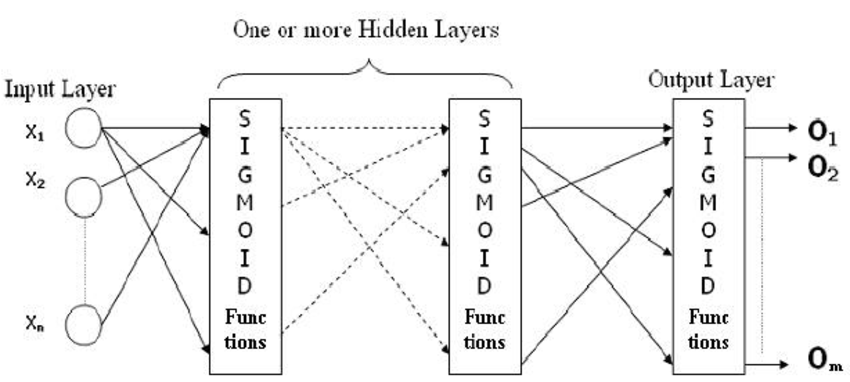}}
	\caption{Multi layer perceptron used in \cite{DBLP:journals/corr/abs-1003-1891}.}
	\label{fig:mlp}
\end{figure}

The MLP classifier used in this method has one input layer, one output layer, and one hidden layer in between, since a single hidden layer suffices for the computation of a uniform approximation of given dataset \cite{DBLP:journals/ker/Kubat99a}. This setting of layers is shown in Fig. \ref{fig:mlp}. Quantity of the neurons residing in each layer are chosen by trial and error. Each of the neurons has sigmoid as the activation function. Supervised training in undertaken for this method, which is performed with the help of 2000 training samples. Back propagation algorithm is used to train the MLP for recognizing the input-output relationship of the problem at hand. 

The results are evaluated with three-fold cross validation, where in each fold the number of neurons are adjusted for optimal performance. By increasing the number of neurons in hidden layer, performance is improved, but only up to a certain threshold point, beyond which increase in number of neurons translates as decrease in performance. This phenomenon is termed as the problem of overfitting \cite{DBLP:conf/icml/Domingos00}. The number of neurons in the hidden layer is finally fixed at 54, and the performance for numeral recognition is obtained at 93.8\%.


\section{Dataset}

Deep learning is solely depended on the data and hence it needs a large amount of data to function properly. Our model is trained and tested on the CMATERDB 3.3.1 Arabic handwritten digit dataset \cite{web4}. The entire dataset consists of 3000 images, making it a source of 3000 unique samples. We divide the dataset in the same way as Das et al. suggested in their studies \cite{DBLP:journals/corr/abs-1003-1891}, which is 2000 training samples and 1000 test samples. 

The dataset consists of 32x32 pixel RGB bitmap images. The images are prepossessed to convert them into gray scale values. Then the image are inverted in order to enhance their feature. As a result, sample image now consists of white foreground and black background, as oppose to conventional white background and black foreground. Then all the images are normalize in order to reduce computation.  

\begin{figure}[!t]
	\centerline{\includegraphics[width=90mm]{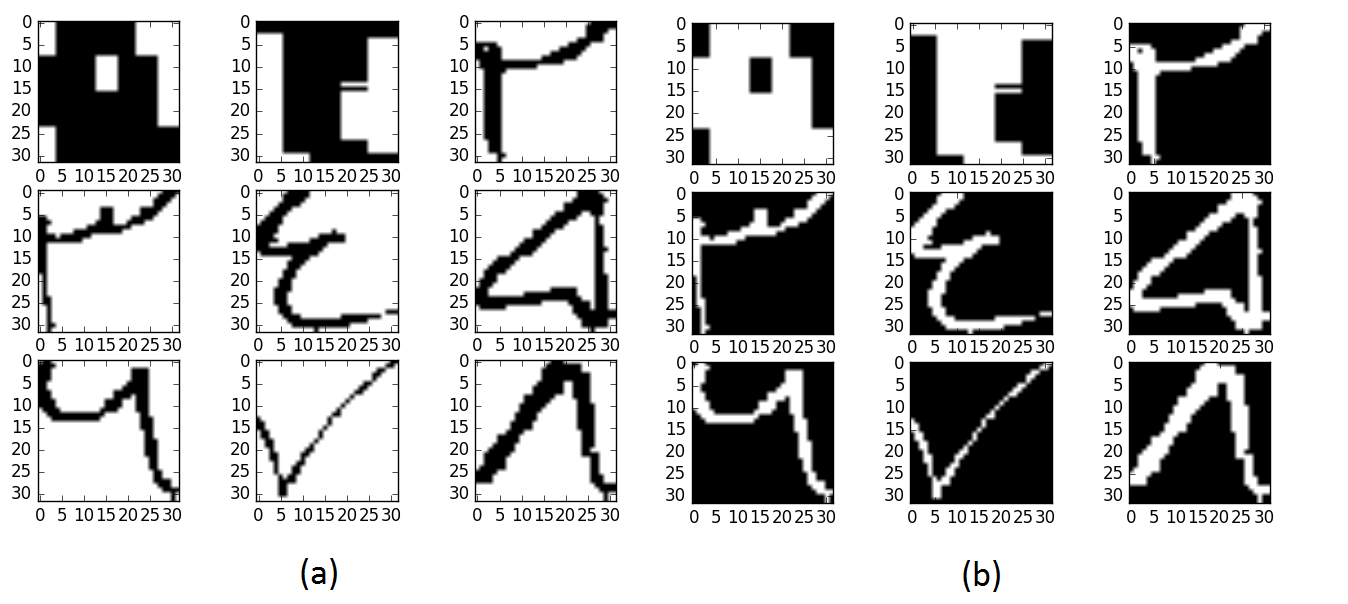}}
	\caption{(a) Raw data and (b) final preprocessed data.}
	\label{fig:raw-v-preprocessed}
\end{figure}

For the MLP model, we reshape the images as a single one-dimensional valued tensor. But for the CNN model, the images are kept in the original dimensions. Fig. \ref{fig:raw-v-preprocessed} shows the sample input image of a digit as opposed to the final prepossessed one.

\section{Proposed Methods}

\begin{figure*}[tb]
	\centering
	\includegraphics[width=1\textwidth]{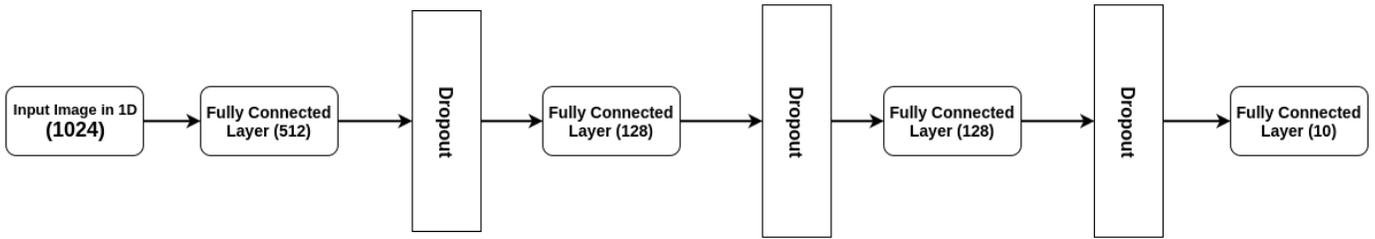}
	\caption{Description of MLP method.}
	\label{fig:arab-inv-model}
\end{figure*}

In this section we discuss our two different methods and how they improve the recognition performance of the given dataset. The first method is the MLP used in the method in \cite{DBLP:journals/corr/abs-1003-1891} with two key changes - we use Rectified Linear Unit (ReLU) as the activation of each neuron in input layer and hidden layer, and use softmax function in the outer classifying layer. In order to tackle the overfitting problem, during training process a number of neurons in each layer are randomly chosen not to get gradient update. This process is called dropout \cite{srivastava2014dropout}. Input layer and intermediate layer have a dropout percentage of 25\% to prevent overfitting. The first output layer of the MLP consists of 512 neurons and take the input image as an one-dimensional array. The intermediate layer consists of 128 neurons and the outer layer consists of 10 neurons. Fig. \ref{fig:arab-inv-model} represents our MLP method. 


\begin{figure*}[!b]
	\centering
	\includegraphics[width=1\textwidth]{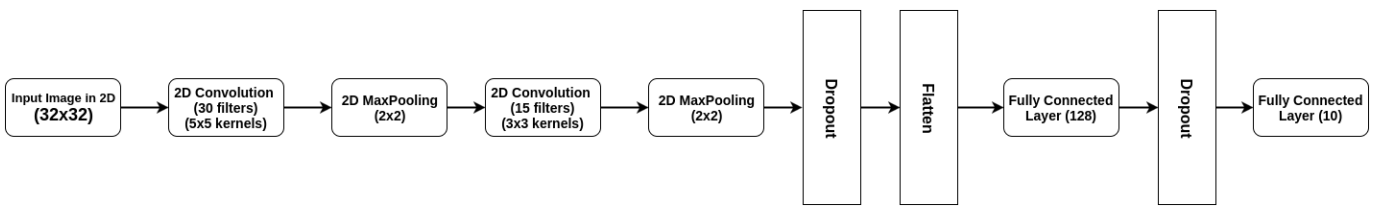}
	\caption{Description of CNN method.}
	\label{fig:arab-inv-model-cnn}
\end{figure*}

Our second method employs Convolutional Neural Network, or Convnets. Convnets are a special variety of Artificial Neural Network (ANN), with trainable weights and biases. Convnets are trained with backpropagation algorithm \cite{hirose1991back}, nonetheless the architecture of the layers is somewhat different \cite{lecun1995convolutional}. 


The first hidden layer is a convolutional layer which has 30 feature maps, each with a kernel size of 5×5 pixels and a ReLU activation function. This is termed as the input layer, which takes in images with 32x32 pixel values. Next we define a pooling layer that takes the maximum value. It is configured with a pool size of 2×2. The following hidden layer is another convolutional layer that has 15 feature maps, each with a kernel size of 3x3 pixels as well as a ReLU activation function. This layer is followed by another pooling layer which is the same as previous pooling layer. 

The next layer is a regularization layer, also called Dropout. It is configured to randomly exclude 25\% of neurons in the layer in order to reduce overfitting. After this we have a layer that converts the two-dimensional matrix data to a vector called Flatten. It allows the output to be processed by standard fully connected layers, which is our next layer. It contains 128 neurons and ReLU activation function, as well as a Dropout layer which is configured to randomly exclude 50\% of neurons in the layer. Finally, the output layer has 10 neurons for the 10 classes or digits in Arabic numeral and a softmax activation function to output probability-like predictions for each class. Fig. \ref{fig:arab-inv-model-cnn} represents our proposed CNN method.

The MLP model takes the whole input image, which has the dimension of 32x32 in a single 1 dimensional array. So the whole image of 1024 pixels are selected as features.  In our convnet approach, we used the same principle. The 2 dimensional convnet takes the whole image as the feature. The MLP method proposed by Das et al. \cite{DBLP:journals/corr/abs-1003-1891} used a pre selected 88 features to feed into the MLP network, however we did not pre-select any features or reduce them. Both of the models takes the whole Image as features.  


\section{Experiment}

Both of the proposed models are trained in the same configuration to evaluate their performances in a comparative way. Both the proposed models, MLP and CNN, are trained and tested against the CMATERDB Arabic handwritten digit dataset and validated against the test data samples. The total parameter size for the proposed MLP method is 591745 whereas for the proposed CNN method the size is 75383. The experiments are done in Keras \cite{chollet2015} along with Theano backend \cite{2016arXiv160502688short}.

Both of the models are trained for 1000 epochs with the batch training size of 128. Adadelta optimizer is used as the optimizing function whereas categorical cross entropy is used to calculate the loss. It is also possible to use mean squared error (MSE) as the loss function; however, that is not conducted in this study. 
All the models are trained with no layer freezed. The supervised training of two proposed methods are done with 2000 training samples, followed by validation with 1000 test samples.

Experimental Models were implemented in Python programming languages with Theano and Keras Library. The experimental setup computer had an intel i5-6200U CPU @ 2.30GHz × 4 with 4 Gb RAM. Nvidia Geforce GTX 625M Graphics card is used to utilize CUDA for faster GPU operation. The whole training process is done for 1000 epoch for each models. Data batch size were 128, during both training and validation.


%

\section{Result and Discussions}

The proposed CNN method outperforms both the proposed MLP method as well as the MLP used in \cite{DBLP:journals/corr/abs-1003-1891}, the latter two yielding identical results for the dataset used in this paper. Table \ref{fig:table} compares the final accuracy of the models.

\begin{table}[!h]
	\centering
	\caption{Performance Comparison of Proposed Methods and Method Described in \cite{DBLP:journals/corr/abs-1003-1891}}
	\begin{small}
		\renewcommand{\arraystretch}{1.5}
		{\begin{tabular}{|M{3cm}|M{2cm}|}
				\hline
				\bfseries Method Name & \bfseries Accuracy \\ \hline
				
				 Das et al. \cite{DBLP:journals/corr/abs-1003-1891} & 93.8 \\ \hline
				Proposed MLP & 93.8 \\ \hline
				proposed CNN & 97.4 \\ \hline
		\end{tabular}}{\label{table_database}}
	\end{small}
\end{table}


After analyzing and evaluating the result of our model for the test samples and validation accuracy, we can come to the conclusion that even thought the MLP model has reached the previous accuracy in \cite{DBLP:journals/corr/abs-1003-1891}, it cannot be improved anymore in the current configuration. 

The computational disadvantage of the basic MLP against CNN here is that, CNN uses the convolution in its layer to detect the features of images in more robust ways \cite{krizhevsky2012imagenet}. The basic CNN model, when trained properly over a dataset, can easily detect image features more accurately, thus performing well above MLP in the fields of classic image recognition problems. In our experiment we demonstrate the performance of the classic MLP. The performance is on par with the performance in \cite{DBLP:journals/corr/abs-1003-1891}. However, the added convolutional layers in the proposed CNN method enhance the accuracy in digit recognition, making the accuracy 97.4 percent. This is the recorded highest accuracy for the CMATERDB Arabic handwritten digit dataset.

\section{Conclusion}

This paper proposes a novel algorithm for recognizing numerals in handwritten Arabic with the help of CNN. Our proposed CNN model uses several convolutional layers along with ReLU activation, with dropout used as a regularization layer. Then the output is fed into a fully connected layer (which is the same as MLP model) with softmax activation to obtain prediction for each class. Our proposed novel method achieves better accuracy then the method discussed in Section \ref{background}. We also propose a modification of the method in Section \ref{background} using MLP. In the MLP model, we implement dropout regularization to reduce overfitting in between fully connected layers. The output layer, consisting of 10 neurons with softmax activation, predicts the probability for 10 individual digit classes (0-9). Our proposed MLP method achieves identical accuracy score compared to the method of Das et al. \cite{DBLP:journals/corr/abs-1003-1891}. The proposed methods are described in detail, and the performances are compared. 

\IEEEtriggeratref{4}
\bibliography{Bib/bib}{}
\bibliographystyle{IEEEtran}


\end{document}